\documentclass[runningheads]{llncs}

\usepackage{eccv}

\usepackage{eccvabbrv}

\usepackage{graphicx}
\usepackage{booktabs}
\usepackage{multirow}
\usepackage[accsupp]{axessibility}  

\usepackage[pagebackref,breaklinks,colorlinks,citecolor=eccvblue]{hyperref}

\usepackage{orcidlink}

\begin{document}

\title{Real-time Transformer-based Open-Vocabulary Detection with Efficient Fusion Head} 

\titlerunning{Abbreviated paper title}

\author{Tiancheng Zhao\inst{1}\orcidlink{0000-0002-4166-6189}  \and
Peng Liu\inst{2} \and
Xuan He\inst{2} \and
Lu Zhang\inst{2} \and 
Kyusong Lee\inst{1}
}

\authorrunning{F.~Author et al.}

\institute{Binjiang Institute of Zhejiang University\\
\email{\{tianchez, kyusongl\}@zju-bj.com}\\
\and
Linker Technology Research\\
\email{\{liu\_peng, he\_xuan, zhang\_lu\}@hzlh.com}}

\maketitle

\begin{abstract}
End-to-end transformer-based detectors (DETRs) have shown exceptional performance in both closed-set and open-vocabulary object detection (OVD) tasks through the integration of language modalities. However, their demanding computational requirements have hindered their practical application in real-time object detection (OD) scenarios. In this paper, we scrutinize the limitations of two leading models in the OVDEval benchmark, OmDet and Grounding-DINO, and introduce OmDet-Turbo. This novel transformer-based real-time OVD model features an innovative Efficient Fusion Head (EFH) module designed to alleviate the bottlenecks observed in OmDet and Grounding-DINO. Notably, OmDet-Turbo-Base  achieves a 100.2 frames per second (FPS) with TensorRT and language cache techniques applied. Notably, in zero-shot scenarios on COCO and LVIS datasets, OmDet-Turbo achieves performance levels nearly on par with current state-of-the-art supervised models. Furthermore, it establishes new state-of-the-art benchmarks on ODinW and OVDEval, boasting an AP of 30.1 and an NMS-AP of 26.86, respectively. The practicality of OmDet-Turbo in industrial applications is underscored by its exceptional performance on benchmark datasets and superior inference speed, positioning it as a compelling choice for real-time object detection tasks. Code: \url{https://github.com/om-ai-lab/OmDet}


\end{abstract}


\section{Introduction}
\label{sec:intro}
Object Detection (OD) is a fundamental task in the field of computer vision that has made significant progress through the integration of various deep neural networks. Traditional close-set OD methods have undergone extensive research and have gradually stabilized, focusing primarily on two directions: improving detector structures to achieve higher accuracy and developing real-time detectors with faster inference speeds. From a stage perspective, well-known methods include the two-stage approach, such as Faster R-CNN\cite{ren2015faster}, and the one-stage approach, such as the YOLO series\cite{redmon2018yolov3, bochkovskiy2020yolov4, long2020pp, glenn_jocher_2022_7347926, ge2021yolox, wang2023yolov7}. From an anchor perspective, research has evolved from anchor-based methods\cite{lin2017focal, wang2023yolov7} to anchor-free methods\cite{tian2019fcos,ge2021yolox,duan2019centernet}. In terms of model structures, two prominent OD architectures are CNN-based detectors and Transformer-based detectors (DETRs)\cite{carion2020end,li2022dn, zhang2022dino, zhu2020deformable, dai2021dynamic, liu2022dab}.

\begin{figure*}[t]
  \centering
  \includegraphics[height=4cm, width=1.0\textwidth]{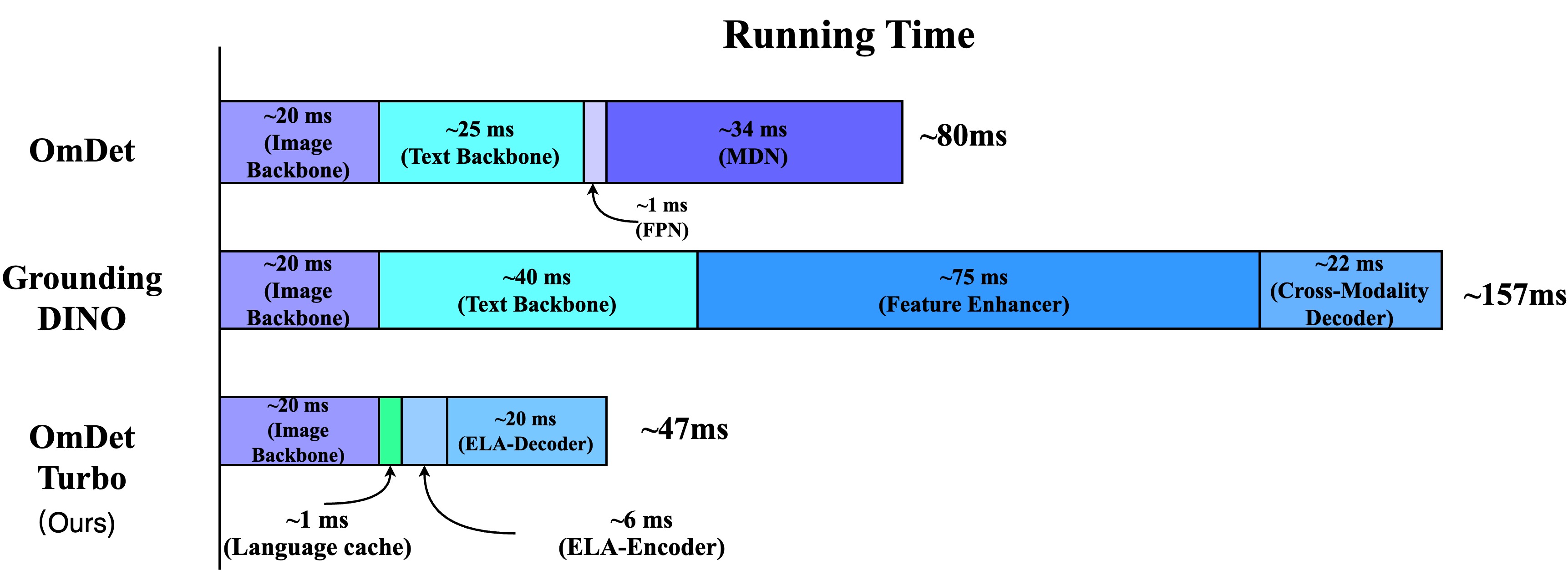}
  \caption{Module-wise speed comparison between the proposed OmDet-Turbo with prior state-of-the-art methods OmDet and Grounding-DINO. (Tested on A100 with PyTorch Implementation.)}
  \label{fig:turbo_model}
\end{figure*}

While most existing methods and real-time detectors are CNN-based detectors, the recent surge of interest in DETRs is notable due to their concise pipeline and end-to-end approach. Several novel DETR methods, such as DINO\cite{zhang2022dino} and Co-DETR\cite{zong2023detrs}, have achieved impressive results on the COCO\cite{lin2014microsoft} dataset, reaching state-of-the-art performance with 63.2 and 65.9 AP, respectively, demonstrating the potential of DETRs. Moreover, RT-DERT\cite{lv2023detrs} achieves real-time object detection surpassing CNN-based detectors through its hybrid encoder. However, traditional OD models share a common limitation—they cannot generalize to objects beyond the training vocabulary. This limitation poses a challenge in meeting the demands of real-world applications and industrial settings.

An emerging research direction in object detection is open-vocabulary object detection (OVD)~\cite{gu2021open,zhou2022detecting}, which aims to detect objects beyond the scope of the training data by incorporating language information to guide the detector. Currently, most OVD models are developed by integrating language modality into close-set detectors. Some OVD models achieve modality fusion by performing early fusion of visual and language information in the encoder or neck components, such as GLIP\cite{li2022grounded}. Some utilize contrastive loss with region features and language features. Additionally, model performance is enhanced through the use of large-scale image-text data and the employment of larger vision backbones and text backbones during training. Among the top-performing OVD models in the OVDEval benchmark\cite{yao2023evaluate}, OmDet\cite{zhao2024omdet} adopts Sparse-RCNN\cite{sun2021sparse} structures and employs Multimodal Detection Network (MDN) to fuse latent queries across recursive heads. On the other hand, Grounding-DINO\cite{liu2023grounding} adopts a DETR structure and enhances its multimodal capabilities by integrating fusion mechanisms across the neck, head, and query initialization stages. Despite these advancements, existing OVD models are encumbered by high computational complexity and prolonged inference times, impeding their practical deployment in commercial applications.

In this paper, we propose OmDet-Turbo, a real-time transformer-based open-vocabulary object detection model.  OmDet-Turbo leverages the DETR structure to maintain a streamlined end-to-end design and achieve robust detection performance. Then through meticulous examination of the structures of OmDet and Grounding-DINO, we identify the bottlenecks in the encoder and ROIAlign~\cite{he2017mask} modules of these models. To overcome these limitations, we propose the Efficient Fusion Head (EFH) that significantly reduces the computation complexity of feature encoding and multimodal fusion, while maintaining good generalization capacity and detection accuracy. Concretely, EFH replaces the heavy encoder in DINO~\cite{zhang2022dino} with a efficient language-aware encoder (ELA-Encoder) that can efficiently predicts prompt-related query proposals given a multi-scale feature map from a visual backbone. Moreover, we introduce the efficient language-aware decoder (ELA-Decoder) that simplifies the vision-language fusion process, enabling multi-task learning and OVD capabilities with fewer fusion components. ELA-Decoder first eliminates the need for slow ROIAlign modules in OmDet with deformable attention mechanism and efficiently fuse features from language and query with simple self attention. We also introduce a decoupled label and prompt embedding structure that allows us to cache text embeddings for faster inference. With these enhancements, OmDet-Turbo combines the robust detection capabilities of DETRs while reducing the computational complexity of resource-intensive modules, resulting in an efficient detection speed suitable for practical deployment.

In addition, we have scaled up the dataset and vocabulary and pre-trained a larger OmDet-Turbo-Base model to demonstrate that our model not only achieves faster inference speed but also attains excellent OVD capabilities with the utilization of a large amount of data. OmDet-Turbo-Base performs competitively with the large versions of other OVD models in terms of zero-shot detection performance on COCO and LVIS\cite{gupta2019lvis} datasets. Furthermore, we have extended our evaluation to two more challenging OVD datasets, ODinW\cite{li2022elevater} and OVDEval\cite{yao2023evaluate}, to explore the model's detection-in-the-wild capabilities and comprehensiveness. OmDet-Turbo-Base achieves state-of-the-art zero-shot performance on these two datasets. This demonstrates that our model maintains a high level of OVD capability while achieving efficient inference speed and leveraging large-scale data. Furthermore, we conduct an ablation study by training a tiny model to demonstrate the efficiency of our model architecture, and performing a time consumption analysis on model components with other models.

The main contributions of this paper are as follows:
\begin{itemize}
\item We propose OmDet-Turbo, a transformer-based real-time open-vocabulary detector that combines strong OVD capabilities with fast inference speed. This model addresses the challenges of efficient detection in open-vocabulary scenarios while maintaining high detection performance.
\item We introduce the Efficient Fusion Head, a swift multimodal fusion module designed to alleviate the computational burden on the encoder and reduce the time consumption of the head with ROI. This module plays a pivotal role in maintaining exceptional OVD performance while enhancing efficiency.
\item Our OmDet-Turbo-Base model, trained on a large-scale dataset, demonstrates excellent zero-shot detection capabilities. It achieves state-of-the-art zero-shot performance on the ODinW and OVDEval datasets, with AP scores of 30.1 and 26.86, respectively. Moreover, the inference speed of OmDet-Turbo-Base on the COCO val2017 dataset can reach 100.2 FPS on an A100 GPU, showcasing its ability to perform real-time open-vocabulary object detection tasks efficiently.
\end{itemize}

\section{Related Work}
\label{sec:formatting}
\subsection{Transformer-based Detection}
Transformer-based object detection methods have gained significant attention in recent years. These approaches leverage the power of transformers to capture long-range dependencies and contextual information in visual data. DETR~\cite{carion2020end} is one of the representatives in this type of model. It uses a transformer encoder-decoder architecture and a set-based global loss that forces unique predictions via bipartite matching. After the release of the DETR model, researchers spent a long period of time optimizing the DETR-like model structure in various aspects. Recent advancements in vision-and-language models like Sparse-RCNN~\cite{sun2021sparse}, DN-DETR~\cite{li2022dn}, DINO~\cite{zhang2022dino}, and RT-DETR~\cite{lv2023detrs} have shown promise in advancing object detection and improving transferability to downstream detection tasks. However, these DETR-like models are traditional closed-set object detection algorithms, which greatly limit the application scenarios and the potential that Transformer should possess.

\subsection{Open-Vocabulary Object Detection}
Open-vocabulary object detection (OVD) is different from traditional object detection algorithms in that users can identify objects using target categories defined in natural language, rather than being limited to pre-defined target categories. Traditional object detection systems are trained on datasets with fixed sets of object classes, limiting their generalization ability. The early OVD algorithm, such as OVR-CNN~\cite{zareian2021open}, trained on fixed-category bounding box annotations and a dataset consisting of image-caption pairs covering a larger variety of different categories. However, due to limited datasets and vocabulary, the generalization performance of such algorithms in open-domain recognition still needs improvement. To address this, recent research has explored vision-and-language approaches, such as CLIP~\cite{radford2021learning} and ALIGN~\cite{jia2021scaling}, which perform cross-modal contrastive learning on large-scale image-text pairs. ViLD~\cite{gu2021zero}, inspired by the CLIP and ALIGN, has built a two-stage object detection model that demonstrates outstanding zero-shot object recognition capability. Another line of work in OVD focuses on building open-world object proposal modules capable of proposing novel objects at test time. GLIP~\cite{li2022grounded} offers a solution by reformulating object detection as a phrase grounding problem, enabling the use of grounding and massive image-text paired data. OmDet~\cite{zhao2022omdet}, inspired by Sparse-RCNN~\cite{sun2021sparse}, treat natural language as a unified way of expressing knowledge. It has a specially designed multimodal detection network (MDN) to solve the problem of manual label taxonomy merging when training with multiple datasets. The following Grounding DINO~\cite{liu2023grounding} combines the transformer-based detection algorithm DINO~\cite{zhang2022dino} with grounding task pre-training, supporting user input of target expressions with attributes, greatly expands the practicality of the OVD model.

\begin{figure*}[t]
  \centering
  \includegraphics[height=6cm, width=1.0\textwidth]{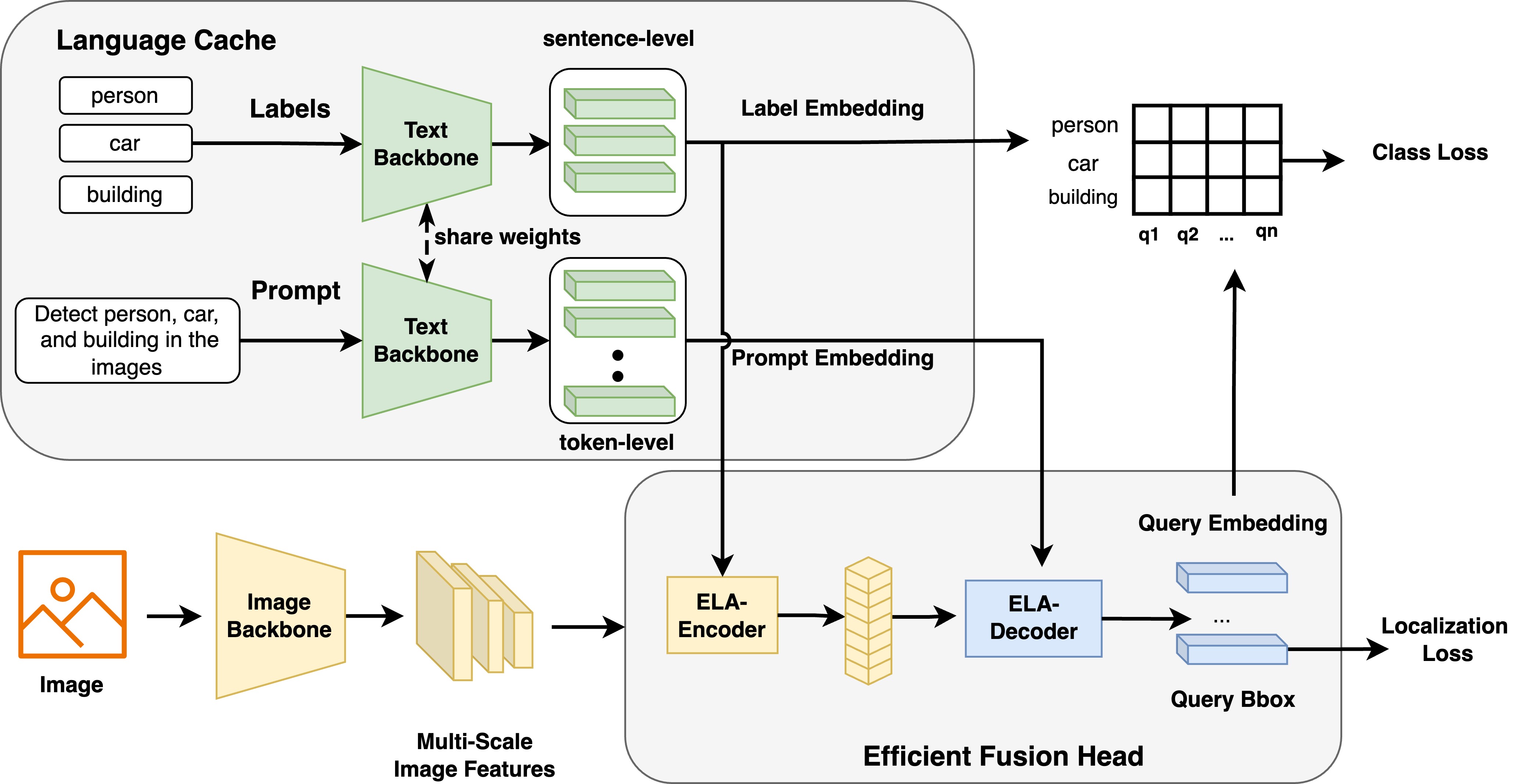}
  \caption{Model Architecture of OmDet-Turbo.}
  \label{fig:turbo_model}
\end{figure*}

Recent researches, such as CORA~\cite{wu2023cora} and BARON~\cite{wu2023aligning}, have endeavored to address certain lingering challenges within the current OVD methods. CORA has sought to employ the multimodal prowess of CLIP in constructing an DETR-style model. Through the utilization of region prompting, CORA strives to bridge the gap between the whole-to-region distribution within the training data. Additionally, CORA employs anchor pre-matching to tackle the task of localizing unannotated targets. BARON attempted to align the region embeddings of the collection of interrelated regions, rather than aligning them individually with the corresponding features extracted from the VLMs.

The limitations of the OVD model lie in its model scale and complexity, making real-time inference difficult. This poses a challenge for widespread adoption in practical applications, which is an issue that needs urgent resolution.

\subsection{Real-time Object Detection}
Object detection models like YOLO series~\cite{bochkovskiy2020yolov4,ge2021yolox,wang2023yolov7}, EfficientDet~\cite{tan2020efficientdet} and RT-DETR~\cite{lv2023detrs} have fast inference speed because they use a one-stage model architecture, which greatly reduces the computational complexity during the inference process. YOLO-World~\cite{cheng2024yolo} is an attempt at real-time open-vocabulary object detection. It inherits the signature one-stage model structure of the YOLO algorithms, uses the clip text encoder to embed the input text, and then uses a specially designed re-parameterizable Vision-Language Path Aggregation Network to fuse text features with image features. YOLO-World has achieved the goal of real-time recognition, but it is a CNN-based method and its generalization accuracy lags behind transformer-based models. 

This paper presents the first real-time transformer-based end-to-end OVD method which achieve strong performance and efficiency on open-vocabulary object detection tasks.

\section{The Proposed Method: OmDet-Turbo}
\label{sec:method}
\subsection{Model Structure}

The model structure of OmDet-Turbo comprises a text backbone $\mathcal{T}$, an image backbone $\mathcal{I}$, and an Efficient Fusion Head (EFH) module. The model's input consists of a prompt describing tasks, a set of object labels, and an image to be detected. To enhance flexibility and performance, we decouple the prompt and label encoders, ensuring they do not share the same text embedding, unlike in GLIP. Labels typically represent the names of objects to be detected, while prompts can vary in nature, such as a combination of object names, a question from a Visual Question Answering (VQA) task, or a generic directive like "Detect all objects" in scenarios of large-vocabulary detection.

The text backbone $\mathcal{T}$ is a transformer language model e.g. CLIP~\cite{radford2021learning} that encodes the text input of the prompt $p$ and K labels $L=[l_1, ..., l_K]$, generating label embeddings and prompt embeddings. Specifically, we use the [cls] tokens from the text backbone outputs as the label embeddings to create sentence-level embeddings of each label $e(l)$ in $L$. For the prompt embeddings $e(p)$, we utilize the token-level embedding outputs from text backbone $\mathcal{T}$ instead of the sentence-level embedding to maintain fine-grained information about the prompt.  The image backbone $\mathcal{I}$ receives the pixel data of the input image and extracts multi-scale image feature pyramid $\{P3, P4, P5\}$. 

We then introduce EFH that is comprised of two key components: Efficient Language-Aware Encoder (ELA-Encoder) and Efficient Language-Aware Decoder (ELA-Decoder). First, the image and text (prompt \& label) features are processed by the ELA-Encoder, which selects top-K initial queries for the deocder. These selected object queries are fused with the prompt embeddings to produce language-guided multi-modality queries within our ELA-decoder. Finally, the output queries from the last decoder layer are used to predict bounding boxes and classes, leveraging the label embeddings $e(l)$ for classification.

\textbf{ELA-Encoder} 
Following the efficient hybrid encoder introduce in~\cite{lv2023detrs}, the last layer of multi-scale image backbone feature $P5$ is encoded via a multi-layer Multi-head Self-Attention (MHSA) module to obtain $F5$. Then Cross-scale Feature-fusion Module (CCFM)~\cite{lv2023detrs} utilizes layers of convolutional layers to fuse the top features from $F5$ to $P4$, $P3$, similar to PANet~\cite{liu2018path}. The encoder output is computed as:
\begin{align}
    \textbf{Q} &= \textbf{K} = \textbf{V} = \text{Flatten}(P5) \\
    F5 &= \text{Reshape}(\text{MHSA}(\textbf{Q}, \textbf{K}, \textbf{V}) \\
    O &= \text{CCFM}(\{P3, P4, F5\}) 
\end{align}
This approach enhances speed by 35\% while maintaining accuracy levels\cite{lv2023detrs}.

Given $O$ from the above encoder, the top-K encoder features is selected as the initial of object position queries in the decoder~\cite{zhang2022dino}. We introduce label embedding to make the selection process language-aware, i.e. select top-K queries related to the current text labels:
\begin{align}
    b_i & = \text{MLP}(o_i) \\
    \alpha_i &= \max_{j\in [1, K]}{\text{Cosine}(e(l_j), o_i))} \quad o_i \in O \\ 
    B_0  &= topK_{\alpha_i \in [i, M]}(b_i, ..., b_M)
\end{align}
where $b_i$ is predicted bounding boxes at each image feature $o_i$, $\alpha_i$ is relevance score of this image feature based on the cosine similarities between $o$ and $e(l)$. The top-K bounding boxes are selected based on the relevance score. 

\textbf{ELA-Decoder} 
Different from Grounding DINO approach, which sequentially utilizes image cross-attention and text cross-attention to fuse image and text features, we simplify the vision-language fusion process as shown in Figure~\ref{fig:turbo_model}. In each decoder layer, we first concatenate the query features and prompt features, followed by a multi-head self-attention mechanism for interaction. Then query features attend to visual features via deformable attention~\cite{zhu2020deformable}, concretely for decoder layer $l$:
\begin{align}
    [Q_{l'}, P_{l+1}] &= \text{MHSA}([Q_{l}, P_{l}]) \\
    Q_{l+1} &= \text{DeformableAttn}(Q_{l'}, B_{l}, O) \\ 
    B_{l+1} & = \text{MLP}(Q_{l+1})
\end{align}
where $Q$ is the query feature, $P$ is the prompt feature, $O$ is the visual feature from ELA-Encoder and $B$ is the position boxes. $Q_0$ is initialized randomly, $P_0$ is initialized from $e(p)$ from the text backbone and $B_{0}$ is initialized from the top-K queries from ELA-Encoder. We further speed up the MHSA component via the Flash Attention method~\cite{dao2023flashattention}. Further during the training process, we employ masking to ensure that the denoising queries in the query features do not interact with other features, including the text prompt features. 


\textbf{Decoupled Prompt and Label Encoder}
Unlike prior grounding-based OVD methods, our approach differs in that we don't directly concatenate all classes as the prompt for object detection. Instead, we encode the prompt of detection task and the classes separately.
By decoupling the prompt and the labels, the prompt of detection task becomes more flexible. A more flexible prompt enables us to easily perform techniques such as language cache and multi-task learning. Furthermore, this approach enables training on datasets with larger vocabulary sizes. Using the concatenated object labels as a prompt could lead to excessively long prompts in scenarios with extensive vocabularies, potentially surpassing the context length limitations of the text backbone. Decoupling the prompt and labels mitigates this issue, ensuring efficient processing and training on datasets with diverse and extensive vocabularies.

\textbf{Language Cache}
In our approach, since we do not perform multi-modal feature interaction during visual and textual feature extraction, the image backbone and text backbone are completely independent. This allows us to extract the text embedding of the target labels and the target detection prompt in advance during testing and deployment phases, storing them in the memory or GPU memory. By doing so, we can avoid redundant embedding extraction for the same text, reducing time consumption during inference. In the training process, if the backbone of the language component is frozen, this method can also reduce significant time expenditure.

\subsection{Model Training}
\textbf{Multi-task Learning}
Previous research on open-vocabulary detection tasks has mostly been limited to detection tasks primarily because these works have not broken free from the fixed mindset of object detection. They simply concatenate the classes to be detected together as the task. By decoupling the classes and the task, it becomes convenient to perform other tasks such as grounding, object detection (OD), visual question answering (VQA), human-object interaction (HOI), etc., and utilize the datasets of these tasks for pre-training. When using grounding datasets, the prompt can involve describing an image, such as ``There is a laptop on the table, next to it is a blue cup.'' When using OD datasets, the prompt can be formed by concatenating the classes together. When using VQA datasets, the prompt can take the form of questions like ``Is there a cup on the table?'' In the case of HOI datasets, the task can involve combinations of people, objects, and relationships, such as ``A bird standing on a person's shoulder.''

\textbf{Scale Up to Large Vocabulary}
The decoupled task, which is independent of the classes, can be applied to large-vocabulary datasets. When there are a large number of target classes to be detected, concatenating them together as the task would result in a very long task. This leads to quadratic increased time consumption when encoding the task in transformer-based language models. Moreover, an excessively long task may even exceed the encoding capacity of the language model. On the other hand, with our decoupled task approach, we use flexible expressions such as ``Detect all objects in the image'' as instructions for the detection model. This allows us to avoid the issue of excessively long tasks and the subsequent exponential growth in encoding time for the language model.

\textbf{Learning Method}
Throughout the training process, in alignment with DINO, we employ multiple denoise groups to expedite model convergence and enhance model precision. Consistent with other detr-base models, during the reconstruction and prediction phases, we employ L1 loss and GIOU loss as the primary loss functions for the detection task. However, in the classification task, after taking the dot product of each query embedding with textual features, we opted not to employ focalloss directly. Instead, we have introduced IoU-aware Query Selection, which is demonstrated to be effective in RT-DETR, to maintain the consistency between the classification and localization of positive samples. Following other detr-based works, an auxiliary loss is added after each decoder layer. The specific computation formula for the loss is as follows:

\begin{align}
\text{Loss}_{\text{od}} & = \lambda{\text{cls}} \cdot L_{\text{cls}} + \lambda_{\text{L1}} \cdot L_{\text{L1}} + \lambda_{\text{giou}} \cdot L_{\text{giou}} \\
\text{Loss}_{\text{dn}} & = \lambda{\text{dn-cls}} \cdot L_{\text{dn-cls}} + \lambda_{\text{dn-L1}} \cdot L_{\text{dn-L1}} + \lambda_{\text{dn-giou}} \cdot L_{\text{dn-giou}} \\
\text{Loss} & = \text{Loss}_{\text{od}} + \text{Loss}_{\text{dn}}
\end{align}


Where $Loss_{od}$ is loss of object detection, $Loss_{dn}$ represents the denoising loss. $L_{cls}$, $L_{dn-cls}$ represents the classification task loss, $L_{L1}$, $L_{giou}$ and $L_{dn-giou}$ are the detection task losses. All the $\lambda$ related parameters are fixed coefficients for each respective component.

\begin{figure*}[t]
  \centering
  \includegraphics[height=6cm]{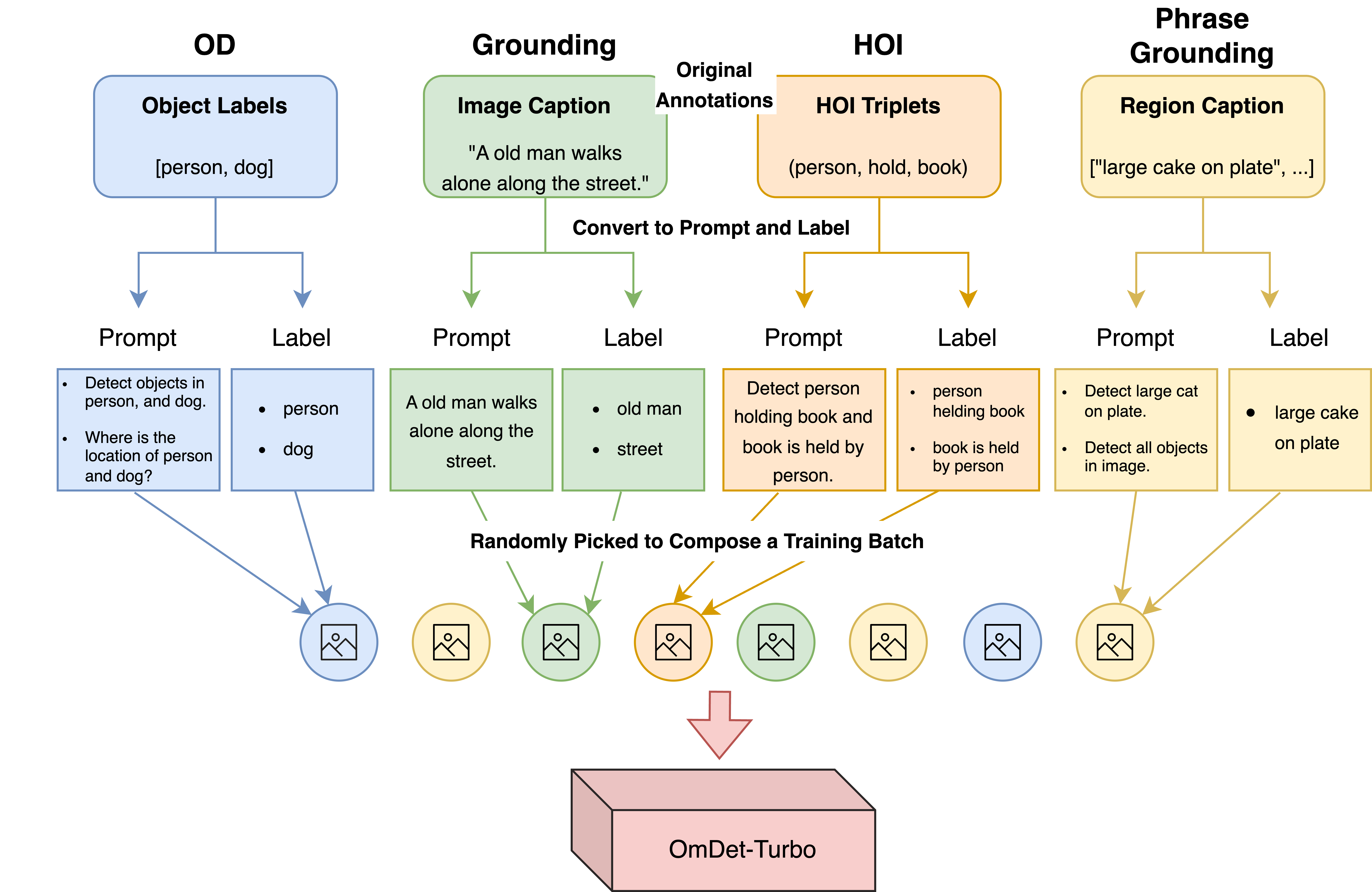}
  \caption{The procedure of our multi-task learning. Initially, the annotated datasets from different tasks are converted into a VQA format. This conversion process involves generating prompts and labels from the original annotations of each task. During training, the converted prompts and labels from various tasks are randomly selected to compose a batch. These prompts and labels are then paired with their respective images. The composed batch, consisting of prompts, labels, and images from different tasks, is fed into the OmDet-Turbo model for training. }
  \label{fig:multitask}
\end{figure*}


\section{Experiments}
\label{sec:exp}
In this section, we conduct experiments using a large-scale dataset to pre-train the OmDet-Turbo-Base model. We evaluate its OVD detection performance and inference efficiency and compare it with other state-of-the-art OVD models that have shown promising results. Furthermore, we train an OmDet-Turbo-Tiny model on identical pre-training datasets and image backbone as other models. This ablation study serves to demonstrate the efficiency of our model architecture and conducting a detailed analysis of time consumption.

\subsection{Experimental Setup}
\textbf{Pre-training data} The pre-training process of OmDet-Turbo-Base involves utilizing datasets from various computer vision tasks through our multi-task learning approach. To enhance the localization ability, we incorporate the O365\cite{shao2019objects365} dataset for object detection. For the grounding task, we select the GoldG\cite{kamath2021mdetr} dataset, as previous studies\cite{li2022grounded, zhao2024omdet} have demonstrated its effectiveness in improving generalization. To improve the model's human-object interaction(HOI) capabilities, we leverage the Hake\cite{li2022hake} dataset and a refined version of HOI-A\cite{liao2020ppdm}. In the refined version of HOI-A, we remove incorrect annotations which used some unreasonable triplet combinations. Furthermore, we incorporate the PhraseCut\cite{wu2020phrasecut} dataset, dedicated to phrase grounding, enhancing the alignment between regions and text within the model. We also used a pseudo labeled large vocabulary object detection dataset with 200K images generated from a VG~\cite{krishna2017visual} pre-trained OD model.

To facilitate training with datasets from various computer vision tasks as figure \ref{fig:multitask} demonstrated, we employ multi-task learning by converting them into a VQA format. Specifically:
\begin{itemize}
\item  Object Detection: For object detection tasks, we concatenate a prefix phrase and object labels to form the prompt. The prefix phrase is randomly selected from a list of templates containing statements like "Detect objects in {}" and questions such as "Where is the location of {}." These templates are designed to enhance the prompt's generalization ability.
\item Grounding: The image caption serves as the prompt input, while object phrases are used as label inputs for grounding tasks.
\item HOI: Original triplets (subject, verb, object) are converted into two object phrases in the form of an adjective followed by a subject or object for HOI tasks.
\item Phrase Grounding: Similar to the object detection task, the dataset is treated in a comparable manner, with the prompt "Detect all objects in the image" used when concatenated label phrases are excessively long.
\end{itemize}

\textbf{Implementation Details} For our OmDet-Turbo-Base model, we employ ConvNext\cite{liu2022convnet} Base as the image backbone and utilize the text encoder from CLIP ViT-B/16 as the text backbone. To ensure stability during training, we freeze the first 6 layers of the text backbone and only fine-tune the last 4 layers. During training, we set the base learning rate to 0.0001 and apply a decay of 0.1 at 70\% and 90\% of the total training steps. The model is trained using 16 NVIDIA A100 GPUs, with a batch size of 64. To facilitate the implementation of OmDet-Turbo, we build the model based on the Detectron2 framework.

\textbf{Evaluation Tasks}
To evaluate the OVD ability of our model, we assess its zero-shot performance on four benchmark datasets: COCO, LVIS, ODinW and OVDEval.

COCO~\cite{lin2014microsoft} dataset is a widely used benchmark for object detection tasks. It consists of a large collection of images with 80 object categories, covering a diverse range of common objects and scenes. And We measure the inference speed of all models on COCO val2017 dataset.

LVIS~\cite{gupta2019lvis} is another important dataset for evaluating object detection models. It contains a more extensive vocabulary of object categories, with over 1,200 classes. LVIS focuses on the long-tail distribution of objects, which presents a significant challenge for detecting rare and less frequent categories. We use LVIS minival to test zero-shot performance and compare with other models.

ODinW~\cite{li2022elevater} (Object Detection in the Wild) serves as a challenging benchmark comprising 35 diverse datasets. This benchmark is specifically designed to assess the task-level transfer capability of pre-trained models in real-world deployment scenarios. We employ the entire set of 35 datasets to comprehensively evaluate the model's performance transfer across various tasks, specifically focusing on the zero-shot setting.

OVDEval~\cite{yao2023evaluate} serves as a benchmark specifically designed to assess the comprehensive zero-shot ability of OVD models. It comprises 9 datasets that encompass 6 linguistic aspects. These datasets include challenging examples with hard negatives, which aim to test the models' understanding of both visual and linguistic input. In our study, we adopt the OVDEval setting and evaluate our model using the NMS-AP metric, which provides a more accurate score.

\subsection{Main Results}
\begin{table*}[t]
\centering
\begin{tabular}{@{}c|ccccc|c@{}}
\toprule
Model  & Model Size & COCO       & LVIS minival          & ODinW         & OVDEval        & FPS             \\ \midrule
GLIP-L     &  430M & 51.4          & 29.3          & -             & 18.4          & 3.1           \\
FIBER-B\cite{dou2022coarse}   & 252M  & 49.3          & \textbf{35.8} & 19.5          & 18.0          & 3.1            \\
Grounding-DINO-B & 233 M   & 56.7$^*$          & 25.1          & -             & 25.3           & 5.7            \\
Grounding-DINO-L   & 341M  & \textbf{60.7}$^*$ & 33.9          & 26.1          & -              & -               \\
Detic-B\cite{zhou2022detecting} &  142M & 45.0$^*$            & 26.8          & -             & 16.0          & 10.1           \\
OmDet-B &   240M   &  57.1    &     34.1    &   19.7   &   25.9     & 9.1 \\ \hline
OmDet-Turbo-B (ours)   &  175M      & 53.4  & 34.7       & \textbf{30.1} & \textbf{26.9} & \textbf{18.6(100.2)} \\ \bottomrule
\end{tabular}
\caption{Zero-shot Results of OmDet-Turbo-Base. * represents supervised score, otherwise it's zero-shot. The FPS for all models were estimated on the COCO val2017 dataset using an A100 GPU. Inference speed without brackets was measured using PyTorch, and speed within brackets was measured using TensorRT.}
\label{tab:omdet_turbo_base_score}
\end{table*}

In Table \ref{tab:omdet_turbo_base_score}, we present a comparison between our OmDet-Turbo-Base model and previous OVD models, such as OmDet, GLIP and Grounding-DINO. To ensure a fair comparison, we select pre-trained models with similar sizes to our model. In cases where a pre-trained model of the desired size is not publicly available, we choose a larger pre-trained model that is accessible. To evaluate the inference speed of all models, we utilize the COCO val 2017 dataset, consisting of 5000 images, and perform the evaluation on a single A100 NVIDIA GPU. 

\textbf{Zero-shot Performance on Common OD Benchmarks} 
COCO and LVIS are widely recognized benchmarks for evaluating OD models. In our evaluation, we assess the zero-shot performance of models on these two benchmarks, specifically focusing on their ability to detect common objects. As shown in Table \ref{tab:omdet_turbo_base_score}, OmDet-Turbo-Base achieves remarkable performance on COCO, achieving an impressive 53.4 AP. Notably, it outperforms GLIP-L, despite having a smaller model size. Although it falls slightly behind the performance of Grounding-DINO models, it's important to note that these two Grounding-DINO models are pre-trained on the COCO dataset. Regarding the zero-shot performance on LVIS, OmDet-Turbo-Base surpasses two large-sized models, GLIP-L and Grounding-DINO-L. This result highlights the strength of our model in detecting objects from a large vocabulary, showcasing its capability in handling diverse detection scenarios.

\textbf{Zero-shot Performance on Complex OD Benchmarks}
In addition to the COCO and LVIS benchmarks, we conducted further evaluations of OmDet-Turbo-Base on the ODinW benchmarks to assess OVD capabilities of our model in real-world scenarios. Our model achieved a zero-shot AP of 30.1, surpassing the performance of Grounding-DINO-L and OmDet-B. This result underscores the transfer ability and adaptability of our model to diverse real-world tasks. OVDEval, a meticulously designed benchmark for OVD, is well-suited to evaluate the comprehensive detection abilities of OVD models. OmDet-Turbo-Base exhibited a strong understanding and proficiency in OVD tasks when evaluated on this challenging benchmark, achieving a new State-of-the-Art (SOTA) score on OVDEval with an NMS-AP score of 26.86. This performance highlights the model's prowess in handling complex and diverse detection scenarios.

\begin{table*}[t]
\centering
\begin{tabular}{@{}c|c|c|c|c|c@{}}
\toprule
Model  &  Backbone  & Pre-Training Data      & COCO      & LVIS minival          & FPS           \\ \midrule
OmDet-T   & Swin T & O365,GoldG    & 46.7       & 28.6       &     12.6(31.2)    \\
Grounding-DINO-T & Swin T & O365,GoldG & \textbf{48.4} & 27.4 & 6.3(38.9)          \\
OmDet-Turbo-T (ours) & Swin T & O365,GoldG  & 42.5            & \textbf{30.3}    & \textbf{21.5(140.0)} \\
 \bottomrule
\end{tabular}
\caption{Zero-shot results of OmDet-Turbo-T on COCO and LVIS. All models are pre-trained on Objects365 and GoldG, and Swin-Transformer Tiny is used as the image backbone. Speed measurement is conducted on A100 GPU in terms of frame per second (FPS) and format is PyTorch speed (TensorRT speed).}
\label{tab:omdet_turbo_tiny_score}
\end{table*}

\textbf{Inference Speed} 
To evaluate the inference speed of models on a practical scale, we conducted inference on 5000 images from the COCO val2017 dataset, which consists of 80 categories. All models are tested using a batch size of 1 on an A100 GPU, and the inference speed is measured in Frames Per Second (FPS). In our analysis, we present two types of inference speed for OmDet-Turbo-Base with language cache applied. The first type is purely measured using PyTorch and Float32 to ensure the results are comparable with other models. The second type is estimated by utilizing TensorRT, which reflects the optimal speed achievable in real-world deployment scenarios. This also demonstrates how our EFH module is enhanced by TensorRT optimization.

In comparison to other models, OmDet-Turbo-Base achieves an impressive inference speed of 18.6 FPS in PyTorch and 100.2 FPS in TensorRT. This rate is approximately 20 times faster than competing models, highlighting the efficiency and speed of our model in processing object detection tasks. This improved performance underscores the practical utility and effectiveness of OmDet-Turbo-Base in real-world applications.

\subsection{Ablation Study}
In our ablation study, we aim to validate the efficacy of our model architecture by training OmDet-Turbo-Tiny using the same pre-training datasets as other models. Previous research has indicated that both Grounding-DINO-T and OmDet-T offer pre-trained versions utilizing Objects365 and GoldG, with the Swin-Transformer\cite{liu2021swin} Tiny serving as the image backbone. Following these established settings, we train OmDet-Turbo-Tiny and assess its performance on the COCO and LVIS datasets to enable a fair comparison with Grounding-DINO-T and OmDet-T.

Moreover, we conduct a detailed analysis of the inference time consumption of each component for all three models. This analysis aims to highlight the bottlenecks present in Grounding-DINO and OmDet structures and demonstrate how our approach effectively addresses and improves upon these bottlenecks to enhance inference speed. By identifying and addressing these bottlenecks, we aim to showcase the efficiency and speed enhancements achieved by our model architecture compared to existing approaches.

\textbf{Performance Analysis} 

Table \ref{tab:omdet_turbo_tiny_score} illustrates the impressive performance of OmDet-Turbo-Tiny when compared to other models in zero-shot scenarios, utilizing identical pre-training datasets and image backbones. Notably, OmDet-Turbo-Tiny achieves a competitive zero-shot score on COCO and excels with a highest score of 30.3 AP on LVIS minival, surpassing the performance of the other models. We further investigate the inference speeds of these models using both PyTorch and TensorRT. When inferring under the PyTorch Float32 setting, OmDet-Turbo-Tiny achieves an FPS of 20.1, demonstrating a speed advantage of approximately 2 times over the other models. Leveraging TensorRT and language cache techniques, OmDet-Turbo-Tiny experiences a remarkable 7-fold improvement, achieving an outstanding inference speed of 140.0 FPS. In contrast, OmDet's inference speed only increases from 12.6 to 31.2 FPS with the same optimizations.The significant speed enhancement observed in OmDet-Turbo-Tiny compared to OmDet-T and Grounding-DINO-T highlights the efficiency and speed benefits inherent in our model architecture for OVD tasks, also showcasing the adaptability of our model structure to TensorRT optimization, emphasizing its practical utility and effectiveness in real-world applications.

\begin{table*}[t]
\resizebox{\textwidth}{!}{
\begin{tabular}{c|cccc|c}
\hline
\multirow{2}*{Model} & \multicolumn{5}{c}{Inference Time (ms)}    \\ \cline{2-6}  &  Text Backbone \vline & Image Backbone   \vline   & Encoder/FPN   \vline   &  Decoder/Head    & Total Time \\ 
\hline
OmDet-T   & 25.2 & 20.0    & \textbf{<1}     & 34.3     &  79.5      \\
Grounding-DINO-T & 40.3 & 20.0  & 75.2 & 22.0 &   157.5    \\
OmDet-Turbo-T (ours) & \textbf{<1} & 20.0    & 6.4   & \textbf{20.1}    &   \textbf{46.5}   \\
\hline
\end{tabular}
}
\caption{Analysis of time consumption on model components: A comparative study among OmDet, Grounding-DINO, and OmDet-Turbo using same tiny-sized models showcases the superior efficiency of OmDet-Turbo's model architecture across multiple components.}
\label{tab:omdet_turbo_tiny_speed}
\end{table*}

\textbf{Detailed Inference Speed Analysis} 
To illustrate the enhancements made to our model and the removal of bottlenecks present in OmDet and Grounding-DINO, we segmented the structures of these three models into four main components: text backbone, image backbone, encoder/FPN, and decoder/head. Subsequently, we meticulously measured the elapsed time of each component. As shown in Table \ref{tab:omdet_turbo_tiny_speed}, OmDet-Turbo consistently outperforms the other two models across all components, with particularly notable improvements in the encoder/FPN and decoder/head sections. In the encoder/FPN component, Grounding-DINO suffers from heavy multi-modality computations in its feature enhancer layer, resulting in significant slowdowns. OmDet-Turbo addresses this bottleneck by implementing a hybrid encoder, leading to a roughly 10-fold speedup in the encoder/FPN process compared to Grounding-DINO. Regarding the decoder/head component, the original MDN in OmDet relies on the time-consuming ROIAlign for feature extraction. OmDet-Turbo introduces the ELA-Decoder, eliminating the need for the ROI (region of interest) operation and substantially accelerating the decoder/head component. Furthermore, OmDet-Turbo capitalizes on its model structure and leverages language cache techniques to eliminate the text backbone during inference, resulting in a time savings of approximately 40 ms for OmDet-Turbo-Tiny.

\section{Conclusion}

In conclusion, this paper introduces OmDet-Turbo, a real-time transformer-based open-vocabulary object detection model that excels in both efficiency and performance. By addressing the challenges of open-vocabulary scenarios while maintaining high detection accuracy, OmDet-Turbo stands out as a compelling solution for real-world object detection tasks. The Efficient Fusion Head module plays a crucial role in enhancing the model's efficiency by reducing computational complexity in the encoder and head components, leading to faster inference speeds without compromising detection performance. OmDet-Turbo-Base, trained on a large-scale dataset, showcases exceptional zero-shot detection capabilities, achieving state-of-the-art performance on challenging datasets like ODinW and OVDEval. With an emphasis on practical deployment and real-time applications, OmDet-Turbo offers a balance between robust detection capabilities and efficient inference speeds. The model's ability to achieve high accuracy in open-vocabulary scenarios, coupled with its impressive performance metrics, positions it as a promising choice for industrial object detection tasks. Through a combination of innovative design choices and meticulous optimization, OmDet-Turbo represents a significant advancement in the field of real-time transformer-based object detection.

%
%
\bibliographystyle{splncs04}
\bibliography{main}
\end{document}